\documentclass[11pt]{article}
\usepackage[margin=1in]{geometry}
\usepackage{authblk}
\usepackage{natbib}
\usepackage{graphicx}     
\usepackage{amsmath, amssymb}
\usepackage{hyperref}
\usepackage{caption}
\usepackage{float}
\usepackage{booktabs}
\usepackage{multirow}
\usepackage{algorithm}
\usepackage{algpseudocode}
\usepackage{subcaption}
\usepackage{graphicx} 

\title{\textbf{MatMMFuse: Multi-Modal Fusion Model for Material Property Prediction}}

\author[1]{Abhiroop Bhattacharya}
\author[1]{Sylvain G. Cloutier}

\affil[1]{Department of Electrical Engineering, École de technologie supérieure, Montréal, Canada}
\affil[ ]{\texttt{\{abhiroop.bhattacharya.1@ens.etsmtl.ca, SylvainG.Cloutier@etsmtl.ca\}}}

\date{\today}

\title{MatMMFuse: Multi-Modal Fusion model for Material Property Prediction }

\begin{document}

\maketitle

\begin{abstract}

The recent progress of using graph based encoding of crystal structures for high throughput material property prediction has been quite successful. However, using a single modality model prevents us from exploiting the advantages of an enhanced features space by combining different representations. Specifically, pre-trained Large language models(LLMs) can encode a large amount of knowledge which is beneficial for training of models. Moreover, the graph encoder is able to learn the local features while the text encoder is able to learn global information such as space group and crystal symmetry. In this work, we propose Material Multi-Modal Fusion(\textbf{MatMMFuse}), a fusion based model which uses a multi-head attention mechanism for the combination of structure aware embedding from the Crystal Graph Convolution Network (CGCNN) and text embeddings from the SciBERT model. We train our model in an end-to-end framework using data from the Materials Project Dataset. We show that our proposed model shows an improvement compared to the vanilla CGCNN and SciBERT model for all four key properties- formation energy, band gap, energy above hull and fermi energy. Specifically, we observe an improvement of 40\% compared to the  vanilla CGCNN model and  68\% compared to the SciBERT model for predicting the formation energy per atom. Importantly, we demonstrate the zero shot performance of the trained model on small curated datasets of Perovskites, Chalcogenides and the Jarvis Dataset. The results show that the proposed model exhibits better zero shot performance than the individual plain vanilla CGCNN and SciBERT model. This enables researchers to deploy the model for specialized industrial applications where collection of training data is prohibitively expensive.
\end{abstract}

\section{Introduction}
Machine learning (ML) has been popular as a potent and adaptable technique in the hunt for materials targeting a wide range of applications, especially when a thorough investigation of the materials space is required \citep{schmidt2019recent,chen2020critical}. With the continuous expansion of high-throughput density functional theory(DFT) datasets and the ongoing development of ML algorithms, it is anticipated that the use of ML for materials discovery will increase even more \citep{saal2013materials,draxl2019nomad,jain2013materials}. Historically, structural descriptors that meet rotational and translational invariance had been used for encoding the crystal structures, ranging from Coulomb matrix \citep{faber2015crystal} and atom-centered symmetry functions (ACSFs) to smooth overlap of atomic positions (SOAP) \citep{behler2011atom,de2016comparing}.

First proposed more than 15 years ago, Graph Neural Networks(GNNs) \citep{scarselli2008graph,gori2005new} have drawn more interest lately in material informatics as a way to overcome static descriptor limitations by learning the representations on adaptable graph-based inputs \citep{li2024evaluating}. Such GNNs have been implemented to predict materials in complex systems including surfaces \citep{palizhati2019toward, back2019convolutional} and periodic crystal arrangements \citep{chen2019graph,xie2018crystal}. The GNN models effectively encode and utilize the structure of the lattice. Particularly, the CGCNN model \citep{xie2018crystal} has shown exemplary performance in encoding the structure property relation while handling periodic boundary conditions. However, the graph convolution based models require a large training dataset to learn generalizable structure property mapping. Moreover, the instances of model failure are difficult to understand and interpret \citep{fung2021benchmarking}. Most importantly, GNN models are unable to incorporate global structural information like crystal symmetry, space group number and rotational information.

Large Language Models (LLMs) provide a promising approach for knowledge discovery in materials science due to their generalization and transferability \citep{jablonka202314}. Their success has motivated applications in structure-property relationship discovery, particularly through pre-trained domain-specific language models, which effectively capture latent knowledge from domain-specific literature. SciBERT, which has been trained on a scientific corpus of 3.17 billion tokens has shown remarkable performance across a diverse set of tasks \citep{beltagy2019scibert}. Compared to graph neural network (GNN) models, LLMs are able to incorporate global information such as space group and crystal symmetry. Combining the strength of the GNN based models with LLM models using multi modal data enhances the feature space, enabling the model to prioritize critical features from diverse latent embeddings. While several studies have explored the potential of LLMs to improve generalization, transferability, and few-shot learning, limited research has focused on integrating textual information from natural language with structural-aware learning from GNNs for crystal property prediction. \cite{li2025hybrid} have used embedding concatenation for combining multiple modalities while, \cite{ock2024unimat} have combined the graph structure of crystals with X-ray diffraction patterns for augmenting the structure aware graph embedding with diffraction information. \cite{lee2025cast} applied masked node prediction pretraining strategy to train a multi-modal model using a combination of text tokens and information from lattice neighbors. However, this architecture might result in locally valid but globally inconsistent structures. \cite{das2023crysmmnet} have developed CrysMMNet which uses concatenation to combine multiple modalities. These models have shown that using multi-modal data with fusion models allows the model to leverage the enhanced feature space. Concatenation uses static connections between modalities and the model design does not focus on cross model connections. While, the proposed model uses cross attention which enables the model to focus on long range dependencies across modalities. Moreover, compared to concatenation, cross attention gives clear attention weights that can be interpreted. To the best of our knowledge, this is the first work, which explores a multi-head attention mechanism to combine structure aware and context aware embeddings to improve prediction and zero shot performance for the prediction of material properties for inorganic crystals.

In this work, we propose, \textbf{Mat}erial \textbf{M}ulti-\textbf{M}odal \textbf{Fus}ion(\textbf{MatMMFuse}), a fusion model which uses a multi-head attention based combination of structure aware embedding of the Crystal Graph Convolution Network (CGCNN) \citep{xie2018crystal} and text embeddings of SciBERT \citep{beltagy2019scibert}. Importantly, we train our model in an end-to-end framework using data from the Materials Project Dataset. We show that MatMMFuse performs in line with state of the art models for four key properties- formation energy, band gap and Fermi Energy. We observe an improvement of 35\% and 68\%  respectively compared to the plain vanilla versions of the model for predicting the formation energy per atom. Furthermore, we demonstrate the zero shot performance of the trained model on small curated datasets of Perovskites, Chalcogenides and the Jarvis Dataset. The primary contributions of this paper are:

\begin{itemize} 
\item Introduction of a multi-head cross attention based fusion approach for accurate material property prediction.
\item Efficiently using multi-modal data to combine structure aware and context aware information to combine local and global information.
\item Improved zero shot performance for specialized materials like Perovskites and Chalcogenides. 
\end{itemize}

\section{Proposed Model Architecture}
\begin{figure}
    \centering
    \includegraphics[width=0.9\linewidth]{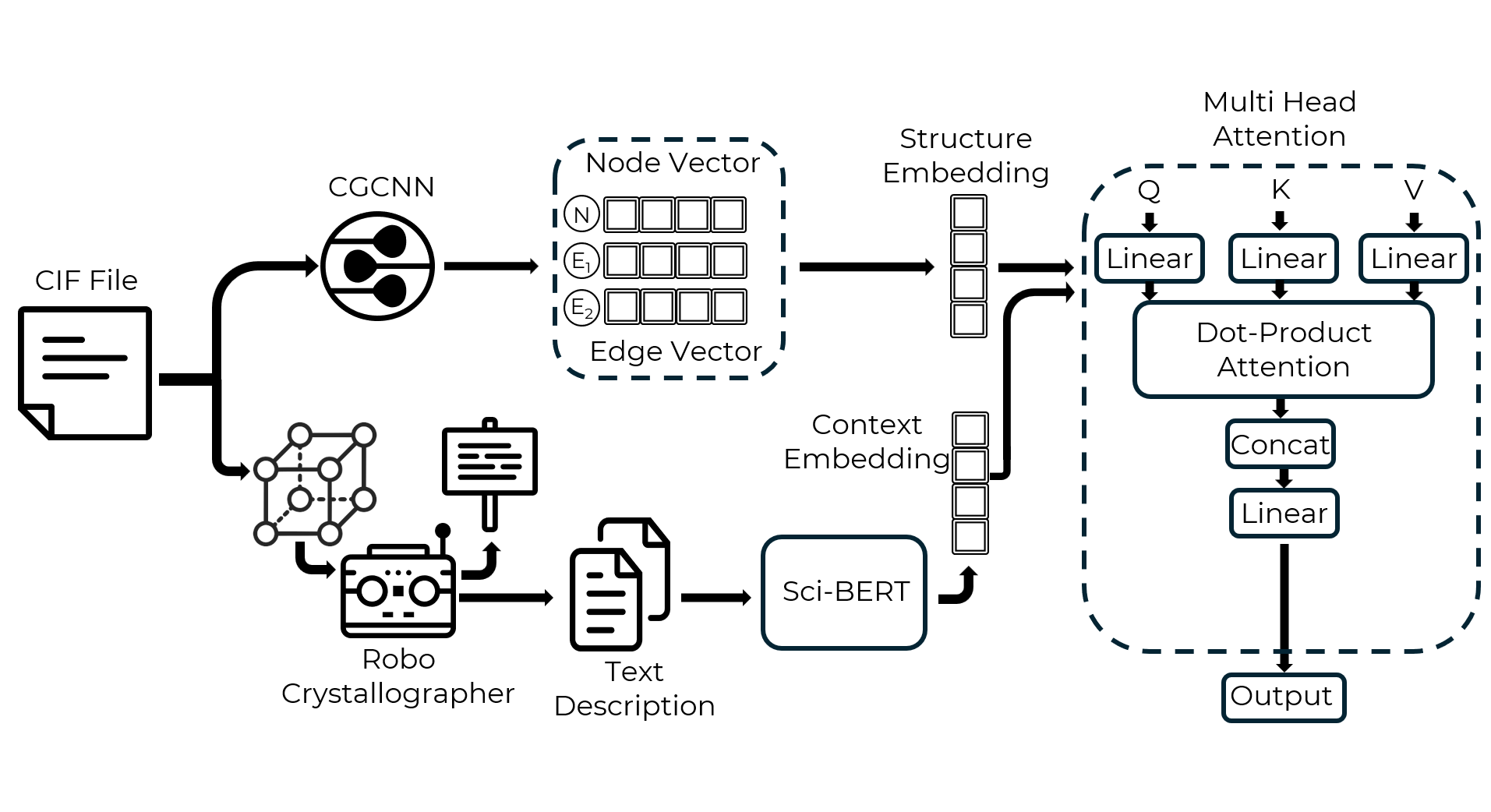}
    \caption{The figure provides an overview of MatMMFuse. The CGCNN model generates a structure aware embedding while the SciBERT model generates a context aware embedding which are combined using a multi-head attention mechanism.}
    \label{fig:schematic_overview}
\end{figure}

The following section describes the architecture of our multi-modal framework. Given a dataset of inorganic crystals denoted by $D =[(S,T),P]$ where $S$, $T$ and $P$ denote the structure information in CIF format, the text description and the material property respectively. The model trains the parameters of the Graph encoder $(G_\theta)$  and the BERT encoder ($B_\theta$) to learn the function $f_\theta \rightarrow P$.  The figure \ref{fig:schematic_overview} captures the model schematic. Each section of the model is explained below.

\subsection{Graph Encoder}
For this model, the material structure from the crystallographic information file(CIF) is encoded as a graph $G(V,E)$  using the CGCNN model where, the atoms are the nodes $V$ and the bonds between the atoms are encoded as the edges $E$. In addition to the graph topology, the node attributes capture the different properties of the atom such as group, position in the periodic table, electro-negativity, first ionization energy, covalent radius, valence electrons, electron affinity and atomic number. For each atom $i$ and it's neighbor $j \in \mathcal{N}(I)$, the convolution updates the atom's feature vector $h_i$ as follows:

\begin{equation}
    h_i^{(l+1)} = h_i^{(l)} + \sum_{j \in \mathcal{N}(i)} \sigma \left( z_{i,j}^{(l)} W_f^{(l)} + b_f^{(l)} \right) \odot g \left( z_{i,j}^{(l)} W_s^{(l)} + b_s^{(l)} \right)
\end{equation}
where, the feature vector of atom i at layer l is denoted by $h_i^{(l+1)}$. The concatenation of $h_i^{(l)}$ and $h_i^{(j)}$ and the edge features $e_{i,j}$ is $z_{i,j}^{(l)}$. $W_f^{(l)}$ and $W_s^{(l)}$ denote the learnable weight matrices. Similarly, $b_f^{(l)}$ and  $b_s^{(l)}$ denote the bias terms. The element wise multiplication is represented by $\odot$ with $\sigma$ and $g$ denoting the activation functions.  
After L graph convolution layers, the graph level representation is obtained by global pooling where in $h_G$ denotes the graph level embedding and $N$ denotes the number of atoms.
\begin{equation} 
    h_G = \frac{1}{N} \sum_{i=1}^N h_i^{(L)}
\end{equation}

\subsection{Text Encoder}
The textual description of the CIF files are generated using the Robocrystallography\citep{ganose2019robocrystallographer} framework. We leverage the scientific knowledge encoded in the pretrained SciBERT model\citep{beltagy2019scibert} followed by a projection layer. For an input sequence $X= (x_1, x_2, \cdots, x_n)$, the self attention mechanism uses the Query Matrix, Key Matrix and Value Matrix denoted by $Q$,$K$ and $V$ respectively. These are linear projections using the corresponding learnable weight matrices.  
\begin{equation}
    \text{Attention}(Q, K, V) = \text{softmax}\left( \frac{Q K^T}{\sqrt{d_k}} \right) V
\end{equation}
It is important to note that BERT uses a multi-head attention mechanism.
\begin{equation}
    \text{MultiHead}(Q, K, V) = \text{Concat}(\text{head}_1, \dots, \text{head}_h) W_O
\end{equation}
A fully connected feed forward network is used with a ReLU activation function and $W_1,W_2$and $b_1,b_2$ learnable weight matrices and biases respectively with the final output obtained by stacking different transformer layers. 
\begin{equation}
    \text{FFN}(x) = \text{ReLU}(x W_1 + b_1) W_2 + b_2
\end{equation}
The model has 12 transformer layers for encoding with 768 hidden dimensions and 12 attention heads. The model has been pre-trained on a 1.14 million papers from Semantic scholar resulting in a total of 3.17 billion tokens.

\subsection{Multi-Head Cross Attention Fusion for Joint Embedding}
The model uses a multi-head cross attention based framework for combining the embeddings generated by the LLM model($h_t$) and the structure aware embedding generated by the GNN (($h_s$). The entire framework is trained in a supervised end-to-end manner. This is a key advantage of the proposed approach because this enables the model to focus on the important sections from the structure aware embedding and the text based embedding.   

\begin{equation}
    Q = W_q h_t, \quad K = W_k h_s, \quad V = W_v h_s
\end{equation}
\begin{equation}
    \text{attention\_scores} = \frac{Q K^T}{\sqrt{d}}
\end{equation}
\begin{equation}
    \text{attention\_weights} = \text{softmax}(\text{attention\_scores})
\end{equation}
\begin{equation}
    \text{combined} = \text{attention\_weights} \cdot V
\end{equation}
The combined embedding is passed through a fully connected layer for the final prediction.
\begin{equation}
   y = W_o \cdot \text{combined} + b_o 
\end{equation}

\section{Experimentation}
We use a Nvidia RTX 4090 graphics processing unit (GPU) to run our experiments. The framework is implemented using the Pytorch library version\citep{paszke2017automatic}. 

\subsection{Dataset}
For model training and assessment, we leverage the widely used Materials Project dataset \citep{jain2013materials}. We focus on four important material properties: the formation energy per atom, the energy above the hull, the fermi energy and the Band Gap. We use 95582 crystal structures with a 80\%,10\%,10\% train, validation and test split. For the CGCNN model, we directly use the data in crystallographic file(CIF) format. We use RoboCrystallographer \citep{ganose2019robocrystallographer} to convert the CIF file to text files. These text files are the input for the SciBert LLM model. The distribution of the target variables and text descriptions are available in the Appendix.
For evaluating the zero shot performance of the model, we use the Cubic Oxide Perovskites, Chalcogenides and a subset of the JARVIS dataset. The distribution of the target variables and text descriptions are available in the Appendix.

\subsection{Experimentation Overview}
We perform experiments in two paradigms. Firstly, \emph{In-domain} wherein we use the traditional approach to train MatMMFuse on examples from the Materials Project dataset. The model is trained in an end-to-end supervised manner. Secondly, we use the trained model to predict the material property of materials with specialized applications without explicitly training on the respective datasets. This paradigm is known as \emph{Zero Shot}. This is intended to be used for specially curated small datasets for materials with specific industrial applications.

\section{Results and Discussions}
\label{sec:results}
\subsection{In-Domain}
We have used MatMMFuse to predict four key material properties. We evaluate the performance of the model for four important material properties - formation energy per atom($E_f$), Fermi Energy($E_g$) and the Band Gap($B_g$). We use AdamW with a cosine learning rate scheduler and warmup. The trained model is then used to predict the formation energy for Perovskites, Chalcogenides and  a subset of the Jarvis Dataset in a zero shot paradigm.  We observe an improvement of 40\% compared to the CGCNN model and 68\% compared to the SciBERT model for the formation energy per atom. However, for the energy above hull, MatMMFuse performs marginally better than SciBERT with a 6.7\% improvement and a 58.5\% improvement over the CGCNN model. The total Fermi energy also shows a similar pattern with a 26\% improvement over the vanilla versions of both the models. Machine learning models have struggled with predictions of the band gap for crystals \citep{zhuo2018predicting} for which the proposed model has an improvement of around 16\% compared to the other models. We hypothesize that the improvement across all the properties is occurring due to the ability of MatMMFuse to selectively combine both local structural information and global information such as space group and symmetry using the attention mechanism.

\begin{table*}[htbp]
\centering
 \caption{Benchmarking model performance. The lower the error the better the model performance.}
\begin{tabular}{@{}ccccc@{}}\toprule
& \multicolumn{4}{c}{Mean Absolute Error($MAE$)} \\
\cmidrule{2-5} 

& \multirow{2}{*}{\shortstack{Formation Energy \\[0.2em] (eV/atom)}} & \multirow{2}{*}{\shortstack{Fermi Energy \\[0.2em] (eV)}} & \multirow{2}{*}{\shortstack{Energy Above Convex Hull \\[0.2em] (eV/atom)}} & \multirow{2}{*}{\shortstack{Band Gap \\[0.2em] (eV)}} \\ 
& & & & \\ \midrule
CGCNN & 0.042 & 0.60 & 0.071 & 0.37\\
SciBERT & 0.081 & 0.59 & 0.031& 0.38 \\
MatMMFuse &\textbf{0.025} & \textbf{0.44} & \textbf{0.029}&\textbf{0.31} \\
\bottomrule
\end{tabular}
\label{tab:benchmarking}
\end{table*}

\begin{figure}
\centering
    \begin{subfigure}[b]{0.40\textwidth}
    \centering
    \includegraphics[scale=0.25]{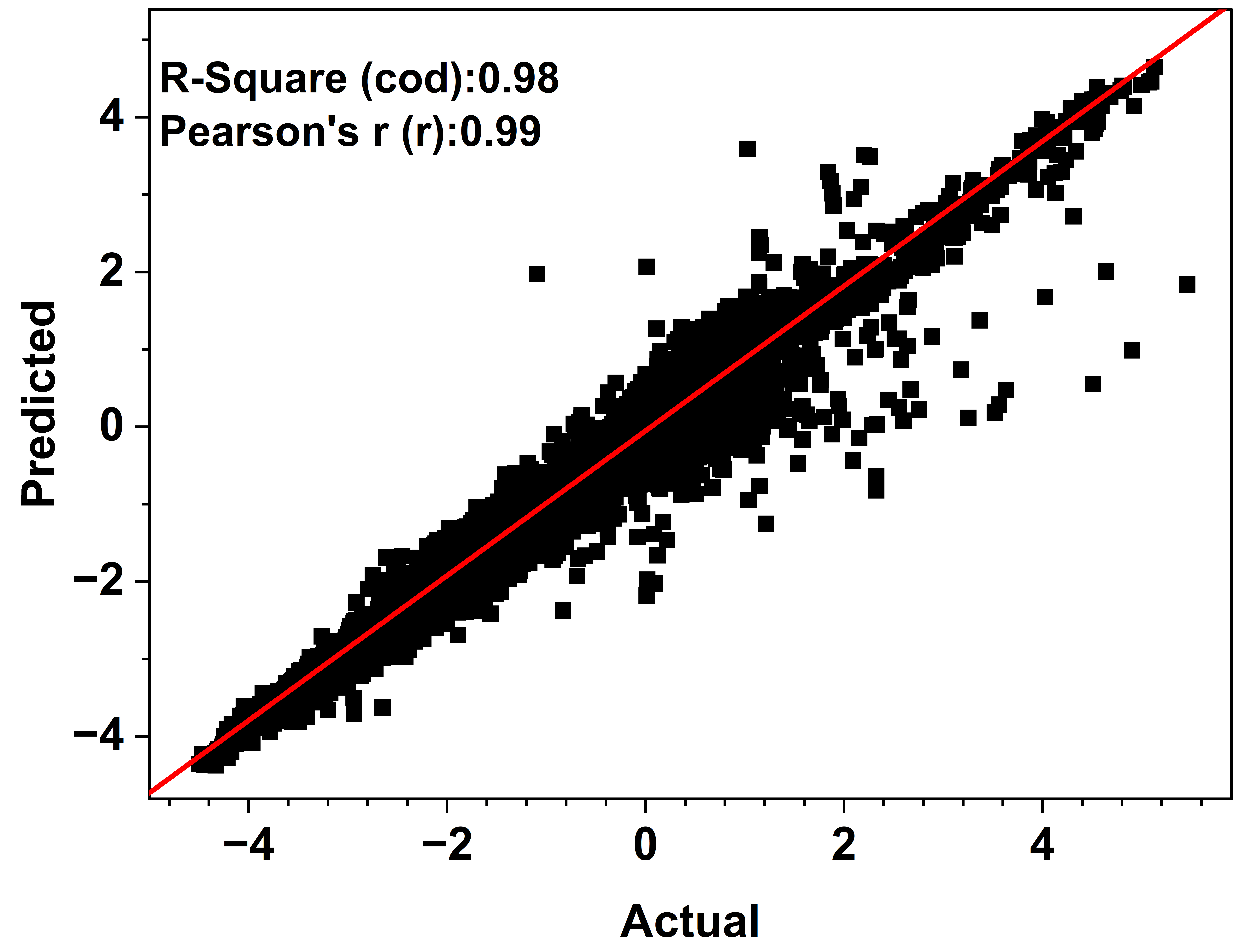}
    \caption{\label{fig:image1} Formation Energy($E_f$)}
    \end{subfigure}
\quad
    \begin{subfigure}[b]{0.40\textwidth}
    \centering
    \includegraphics[scale=0.25]{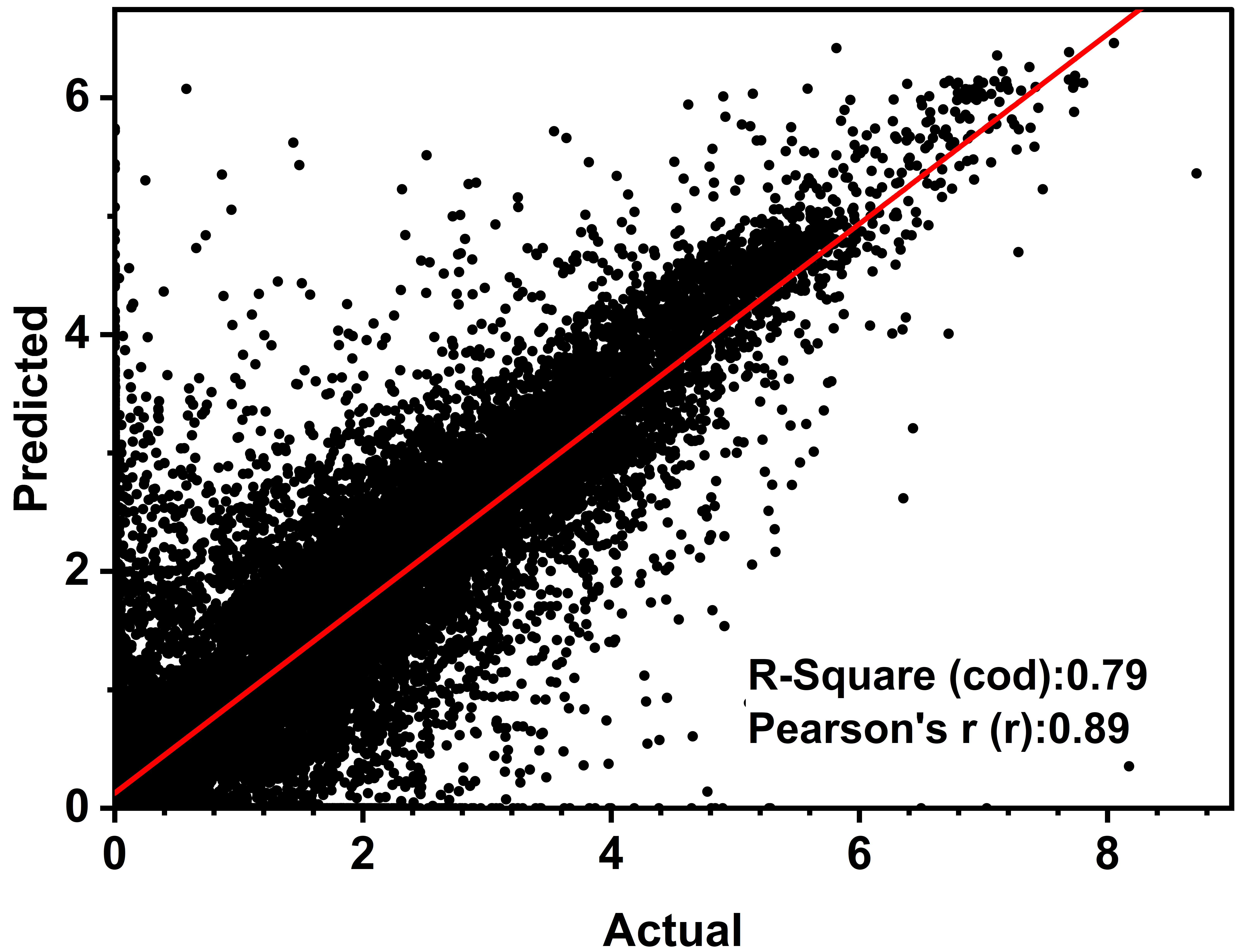}
    \caption{\label{fig:image2} Band Gap($B_g$)}
    \end{subfigure}
\caption{The scatter plot presents the actual versus predicted values for (\subref{fig:image1}) Formation Energy per atom, (\subref{fig:image2}) Band Gap. The model tends to incorrectly predict the values for band gap when the values are close to zero.}
\label{fig:avf_indomain}
\end{figure}

To further investigate the results, we are comparing the plots of the actual versus predicted values for the formation energy per atom and the band gap for the test dataset \ref{fig:avf_indomain}. For formation energy, we observe that the predictions are aligned with the actual values with a $R^2$ of 0.97 while, for Band Gap we can clearly see that the model predicts a higher value when the actual value is close to zero.  

\begin{figure}[htbp]
\centering
    \begin{subfigure}[b]{0.3\textwidth}
    \centering
    \includegraphics[width=5.4cm]{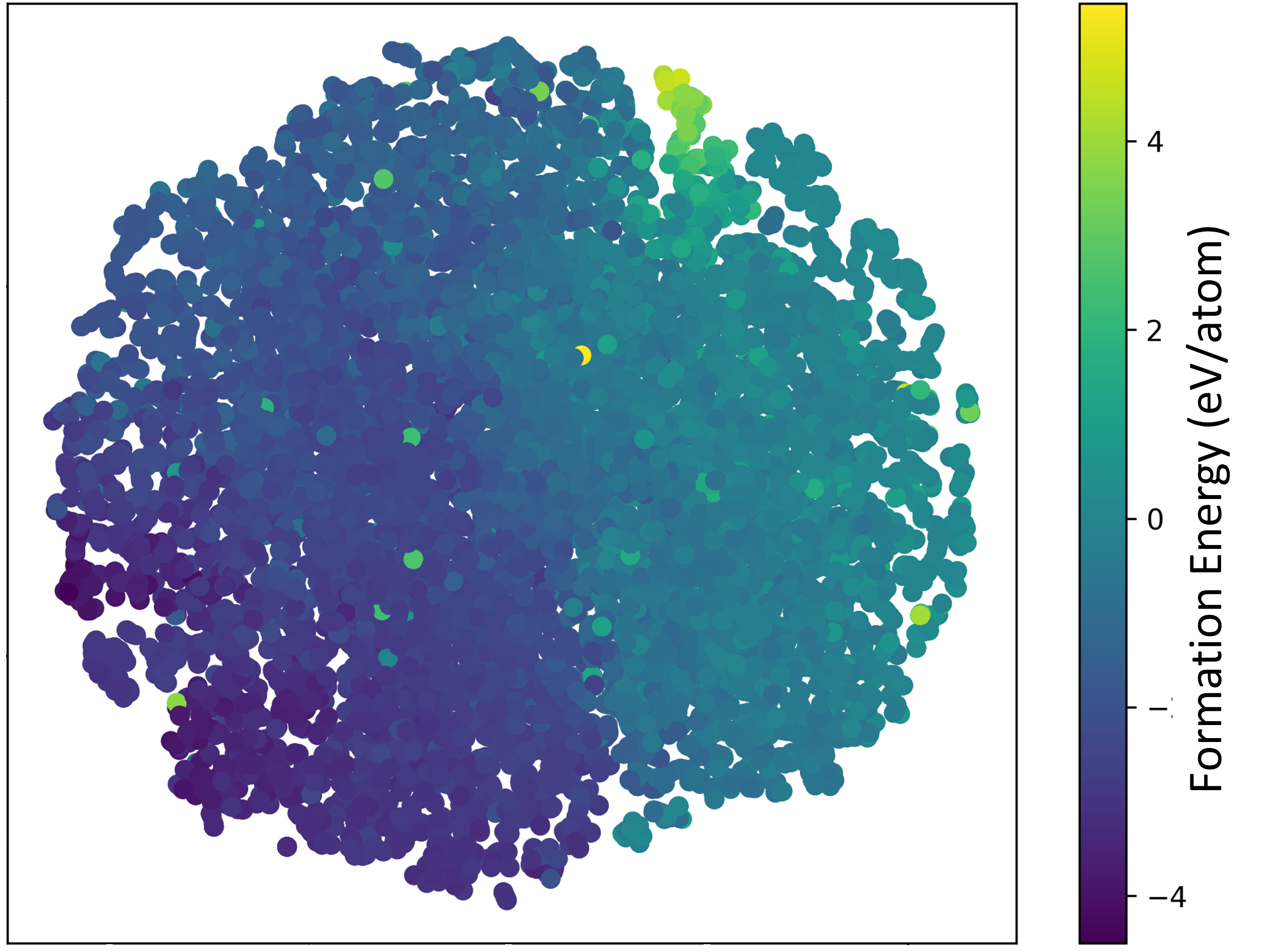}
    \caption{\label{fig:tsne1}}
    \end{subfigure}
\quad
    \begin{subfigure}[b]{0.30\textwidth}
    \centering
    \includegraphics[width=5.4cm]{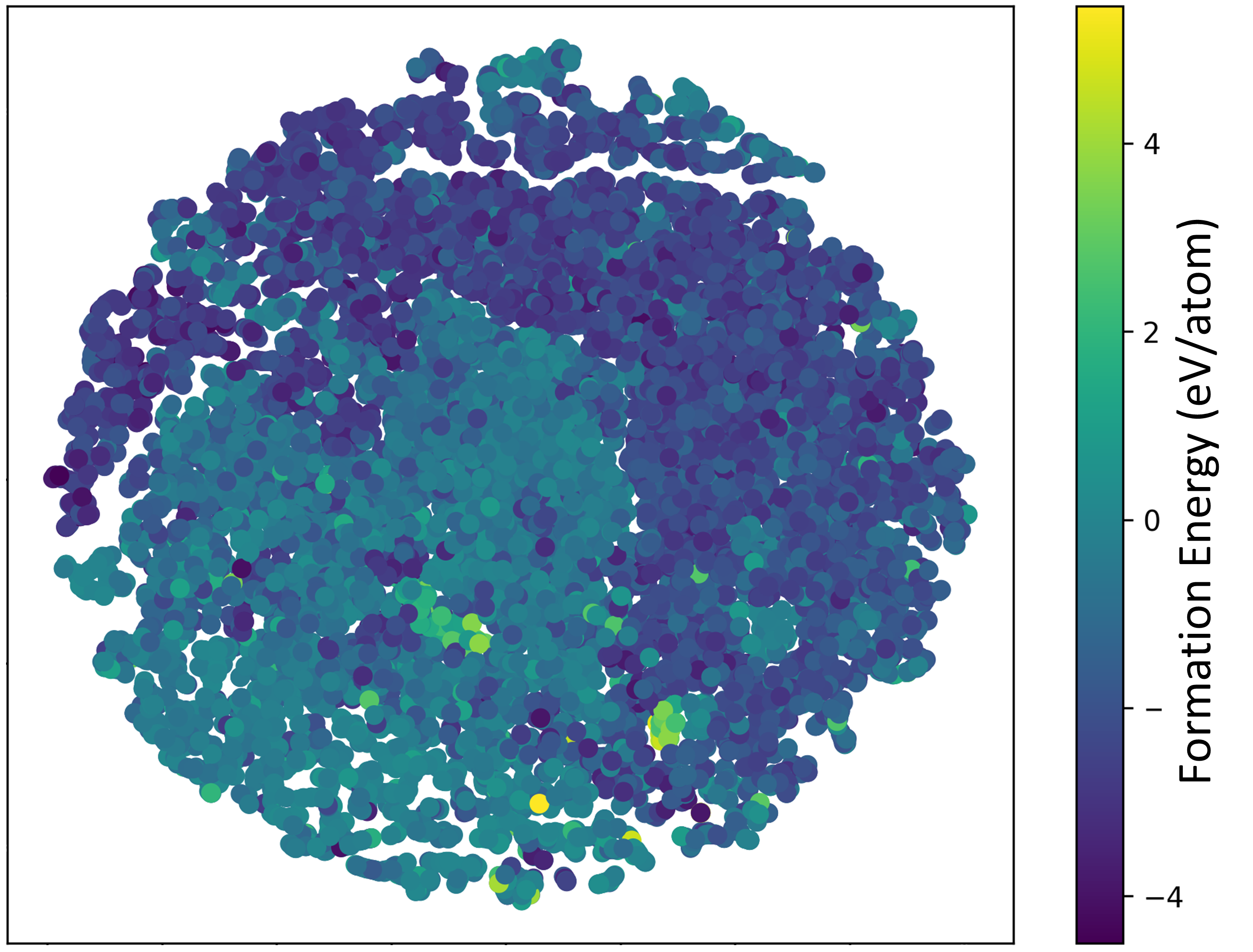}
    \caption{\label{fig:tsne2}}
    \end{subfigure}
\quad
    \begin{subfigure}[b]{0.30\textwidth}
    \centering
    \includegraphics[width=5.4cm]{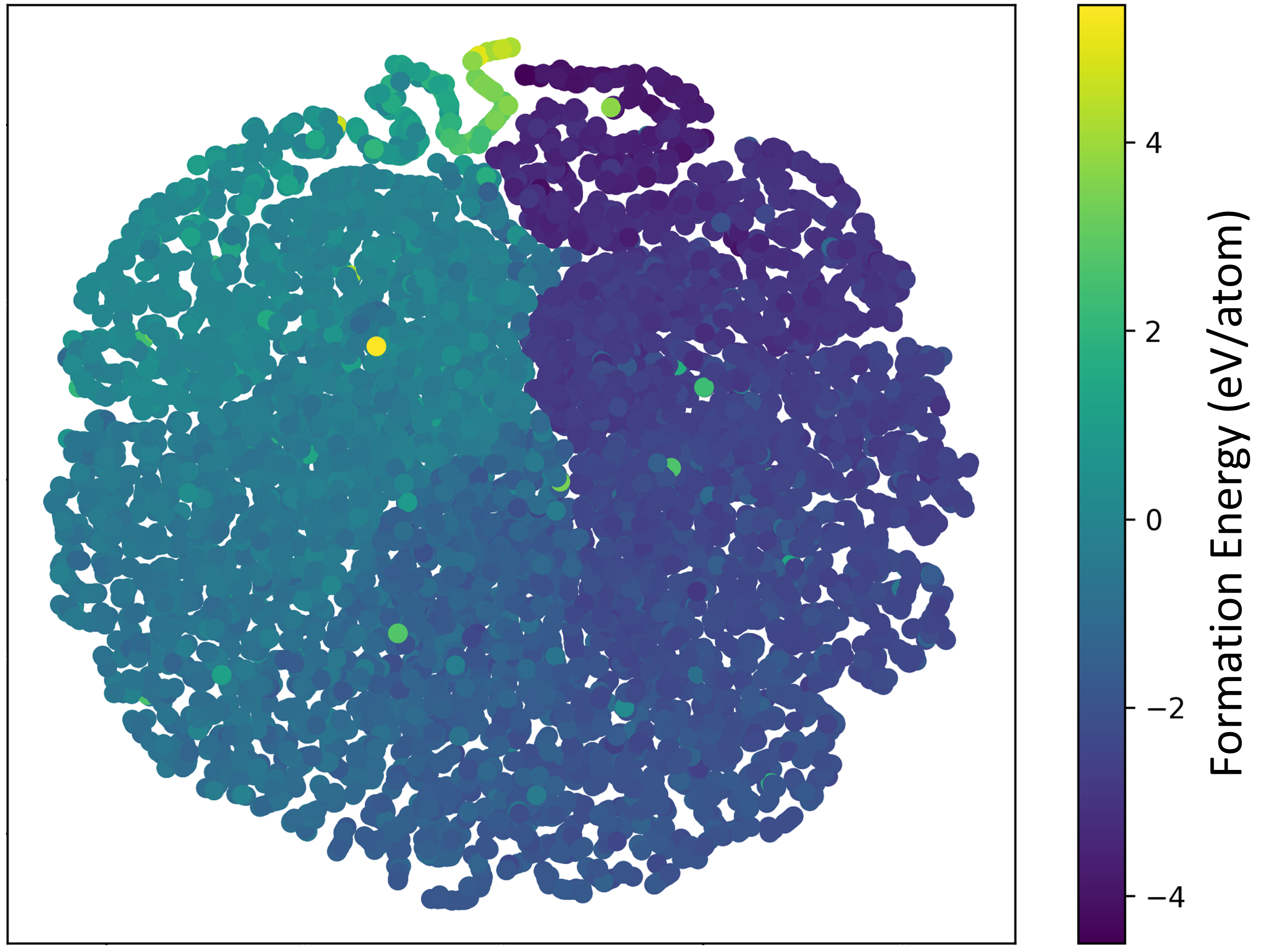}
    \caption{\label{fig:tsne3}}
    \end{subfigure}
\caption{The t-SNE plot of the embedding from the embedding for the test dataset from Materials Project, with each point representing an individual crystal. Colors for each point are associated with their formation energies. (\subref{fig:tsne1}) CGCNN embedding, (\subref{fig:tsne2}) SciBERT embedding , (\subref{fig:tsne3}) MatMMFuse embedding}
\label{fig:tsne}
\end{figure}

The t-distributed stochastic neighbor method (t-SNE) \citep{van2008visualizing} allows us to understand the decision boundaries and segregation of data points in the high dimensional embedding using 2D plots. The Figure \ref{fig:tsne} depicts a combined structure-composition latent space for the trained materials, in which points within a grouping are anticipated to have similarities in both their atomic structures and elemental compositions. We see comparable clustering in the latent space. In the t-SNE plot of MatMMFuse we observe that the dark and light colored ones are segregated in different clusters with lobe-structured decision boundaries which shows that the learned embedding is able to discern between crystals with high formation energies and ones with low formation energy. We observe decision boundaries in the embedding generated by the SciBERT model as well but the points are not clustered. The embedding generated by the graph encoder does not have clear clustering or decision boundaries.

\subsection{Zero Shot Performance}
A key challenge in material science is the lack of large datasets for specialized applications. Most materials with specialized applications such as photovoltaic cells and battery, do not have large datasets with DFT calculated material properties to enable training of data hungry deep learning models. In this section, we demonstrate that the trained MatMMFuse model can be used for predicting the material properties for small curated datasets in a zero shot manner. The proposed attention-based method for combining embeddings leads to an improvement in the zero shot performance of the model. Attention allows the model to dynamically weight and combine embeddings based on the relevance to task enabling the model to focus on the most informative features from each embedding. In the table \ref{tab:zero_shot}, we compare the zero shot performance of MatMMFuse for predicting the energy of Perovskites, Chalcogenides and a small subset of the Jarvis dataset with the vanilla CGCNN and SciBERT models.

\subsubsection{Cubic Oxide Perovskites}
$ABO_3$ perovskites are viewed as promising resistive-type gas sensors \citep{ishihara2009structure}. It is important to remember that there are 2704 observations in
the dataset which is insufficient for training large GNN or LLM models. MatMMFuse achieves a MAE of 1.28 on the test dataset which is 10\% lower than the CGCNN model and 55\% lower than the SciBERT model. 
\subsubsection{Chalcogenide Perovskites}
For photovoltaic applications researchers have proposed Chalcogenide perovskites of the form $AB(S, Se)_3$ because of their
stability, non-toxicity, and lead-free composition \citep{basera2022chalcogenide}. To test our model, we repeated the same experiment for a dataset for $AB(S, Se)_3$ perovskites. The dataset has 1621 observations. Nonetheless, MatMMFuse achieves a low MAE of 1.05, lower by 21\% and 27\% as compared to the CGCNN and SciBERT models respectively.

\subsubsection{JARVIS}
The JARVIS (Joint Automated Repository for Various Integrated Simulations) dataset \citep{choudhary2020joint} is a high-throughput materials database developed by the National Institute of Standards and Technology(NIST). The dataset encompasses a wide array of materials properties, computed using density functional theory (DFT) simulations. MatMMFuse achieves a MAE of 0.078 which is 48\% lower than the CGCNN and the 59\% lower than the SciBERT model.

\begin{figure}
    \centering
    \includegraphics[scale=0.3]{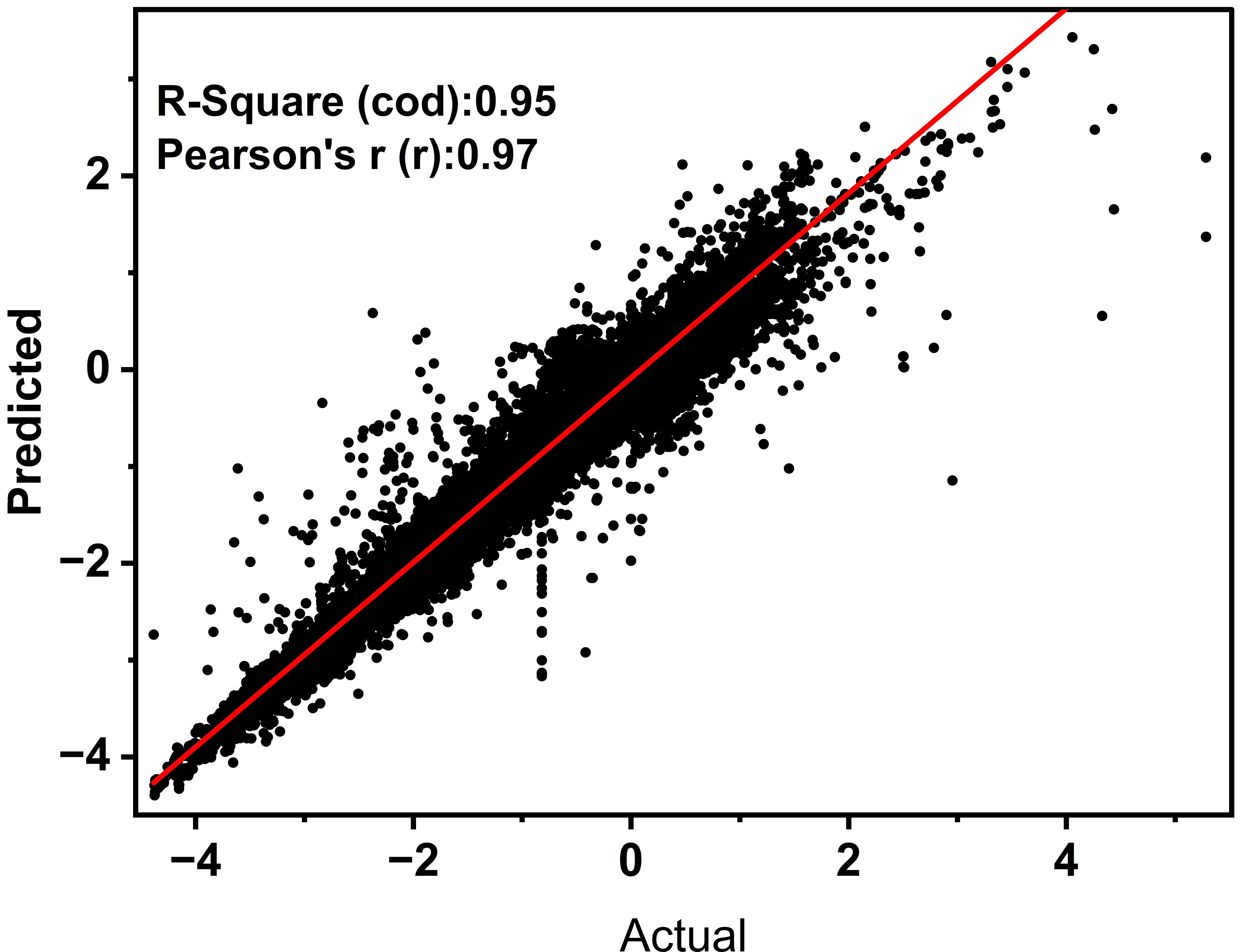}
    \caption{The scatter plot presents a comparison between the actual and the predicted values of formation energy per atom for the JARVIS dataset.}
    \label{fig:jarvis_avf}
\end{figure}

The actual versus predicted curve in Figure\ref{fig:jarvis_avf}  shows that the predicted and actual values are aligned with a $R^2$ of 0.94. However, there are a number of data points around -0.9 eV/atom which have a much lower prediction. 

\begin{table*}[htbp]
\centering
 \caption{Zero Shot performance for the Energy per atom. The lower the error the better the model performance.}
\begin{tabular}{@{}ccccc@{}}\toprule
& \multicolumn{2}{c}{Energy $MAE(eV/atom)$} &\\
\cmidrule{2-4} 
& $CGCNN$ & $SciBERT$ & $Proposed$\\ \midrule
Perovskites($ABO_3$) & 1.42 &2.84  & \textbf{1.28}\\
Chalcogenides($ABS_3$) & 1.33 & 1.44  &\textbf{1.05}\\
JARVIS &0.15 & 0.19 & \textbf{0.08} \\
\bottomrule
\end{tabular}
\label{tab:zero_shot}
\end{table*}

\subsection{Ablation Studies}
This section presents the ablation studies performed by changing, adding or removing the key parts or inputs of the model architecture.

\subsubsection{Encoded domain knowledge}
\begin{figure}
    \centering
    \includegraphics[width=0.9\linewidth]{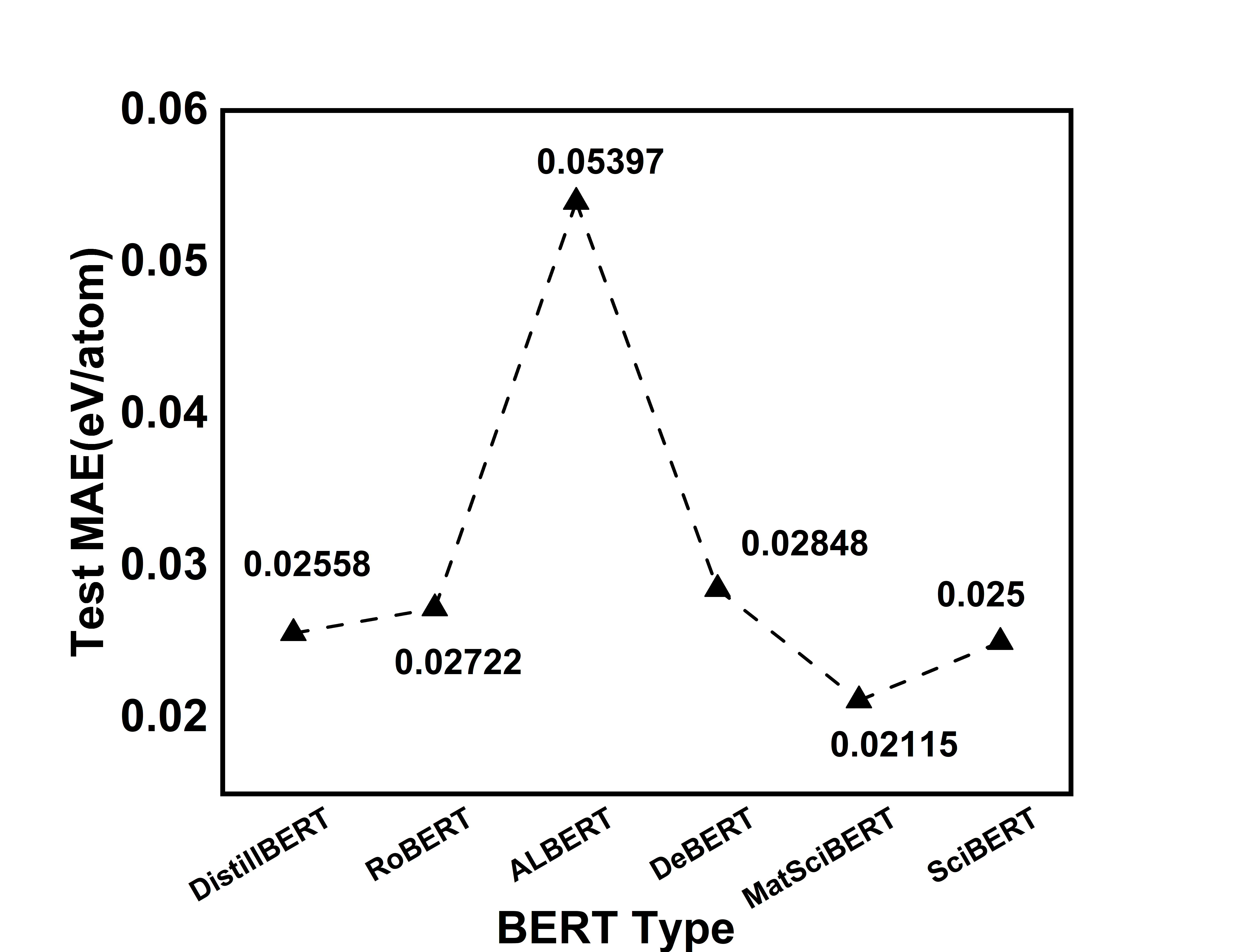}
    \caption{The plot compares the performance of different BERT models for encoding the text representation. MatSciBERT has the best performance and ALBERT has the worst performance. }
    \label{fig:ablation_bert}
\end{figure}

To prove our hypothesis that MatMMFuse is able to leverage the encoded knowledge in the LLM Model, we have run experiments by using variations of the BERT model as the text encoder for predicting the formation energy per atom. The alternate models used are ALBERT \citep{lan2019albert}, RoBERT \citep{masala2020robert}, DeBERT \citep{sergio2021stacked} and DistillBERT \citep{sanh2019distilbert}. Due to the knowledge of material science encoded in the MatSciBERT model, we observe that it outperforms all models closely followed by SciBERT model. It is important to note that ALBERT shows a sharp deterioration in model performance. We posit that this might be due to two reasons. Firstly, ALBERT shares parameters across all transformer layers, reducing model size but limiting the model's capacity to learn distinct representations at different levels of abstraction. Secondly, it uses token order prediction as compared to next token prediction used in other BERT Models. The figure\ref{fig:ablation_bert} presents a comparison of model performance for different BERT models.

Further to this, we also observe a similar improvement in the zero shot performance of the model on the specialized cubic oxide Perovskite, Chalcogenides and the JARVIS dataset which is shown in table\ref{tab:matbet_zero_shot}

\begin{table*}[htbp]
\centering
 \caption{Zero Shot performance of the MatSciBERT model for the Formation Energy per atom. The lower the error the better the model performance.}
\begin{tabular}{@{}ccccc@{}}\toprule
& \multicolumn{2}{c}{$MAE(eV/atom)$} &\\
\cmidrule{2-4} 
& $MATSciBERT$ & $SciBERT$\\ \midrule
Perovskites($ABO_3$) & 2.26 & \textbf{1.28}\\
Chalcogenides($ABS_3$) & \textbf{1.33}  &0.98\\
JARVIS &\textbf{0.037} & 0.08 \\
\bottomrule
\end{tabular}
\label{tab:matbet_zero_shot}
\end{table*}

We have decided to use SciBERT model because it has more interdisciplinary knowledge which leads to a more broader scientific context. Especially, for applications in biomedicine and energy. $ABO_3$ perovskites are used for solar cells and therefore SciBERT outperforms MatsciBERT in such specialized applications.

\subsubsection{Encoded lattice structure}
\begin{figure}
    \centering
    \includegraphics[width=0.9\linewidth]{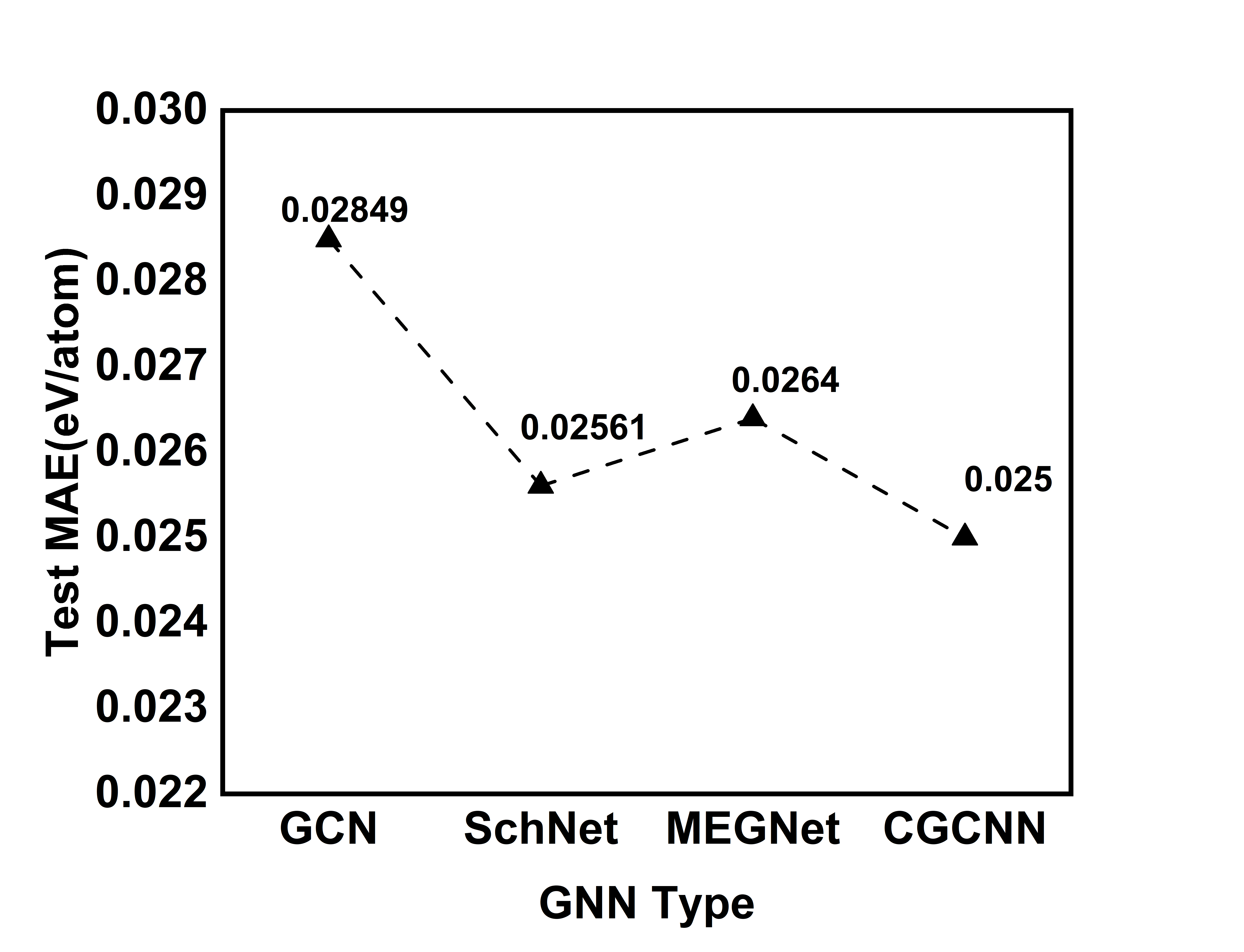}
    \caption{The plot compares the performance of different GNN models for encoding the lattice structure. CGCNN gives the optimum tradeoff between performance and efficiency. }
    \label{fig:ablation_gnn}
\end{figure}

The encoding of the crystal lattice structure using different graph encoding models results in different ways of capturing the complex relationships within crystal structures. We used SchNet \citep{schutt2018schnet}, MEGNet \citep{chen2019graph}, CGCNN and Graph convolution networks(GCN) for the analysis. Vanilla GCN architectures are not designed to incorporate periodic boundary conditions. On the other hand, SchNET and CGCNN explicitly incorporate crystal periodicity. CGCNN uses discretized bins for edge features while SchNet uses continuous radial basis functions for smooth distance representation. CGCNN includes more extensive information about the crystal structure and is computationally more efficient compared to SchNET which uses continuous-filter convolutions with filter-generating networks that create customized filters for each atomic interaction based on distance. Unlike other models, MEGNet uses global state variables such as unit cell parameters which makes it more expressive but also more computationally expensive. We found that CGCNN gives the optimum tradeoff between performance and efficiency. The figure\ref{fig:ablation_gnn} presents a comparison of model performance for different GNN based encoder models.

\subsubsection{Multi-Head Attention Module} 
\begin{figure}
    \centering
    \includegraphics[width=0.9\linewidth]{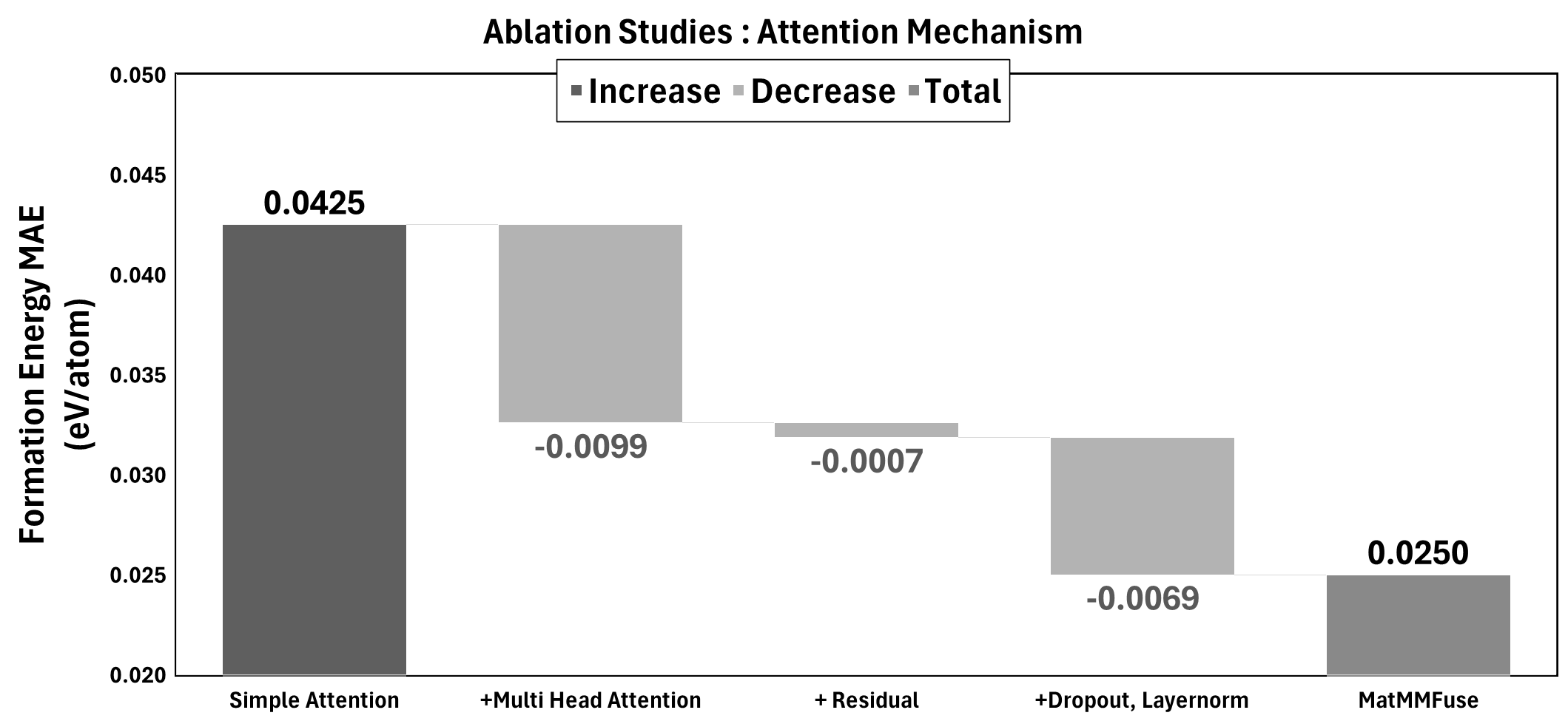}
    \caption{The waterfall chart shows the effect of adding individual components to improve the attention based fusion method.}
    \label{fig:ablation_attn}
\end{figure}

A comprehensive ablation study was performed on the different sub-modules of the attention based fusion mechanism for allowing the model to focus on the specific parts of the structure aware and context aware embeddings. We observe that using a multi-head attention method considerably improves the performance. Using a layer-norm to normalize the layer outputs increases the stability of the training and helps the model converge. For small datasets, including dropout prevents overfitting and helps the model generalize better. Interestingly, including a residual layer does not lead to a significant improvement in model performance.The waterfall chart\ref{fig:ablation_attn} captures the effect of each change on the model performance.

\subsubsection{Robustness to Training data size}
\begin{figure}
     \centering
\includegraphics[width=0.8\linewidth]{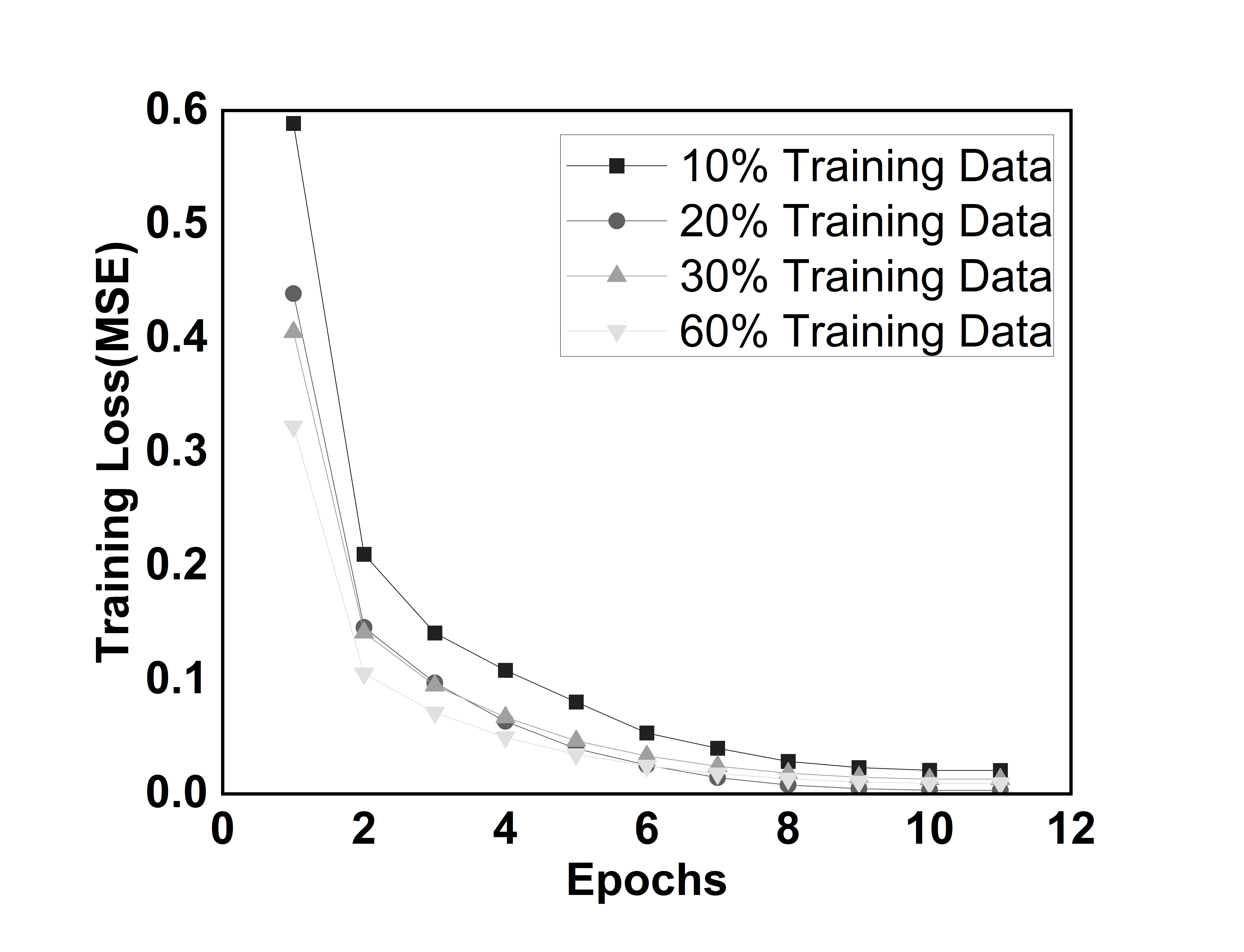}
    \caption{The chart shows that MatMMFuse is relatively robust to reduction in training data.}
    \label{fig:training_data}
\end{figure}

As reported in literature, reducing the size of the training data reduces the performance of the CGCNN model \citep{xie2018crystal}. Interestingly, we observe that using an enhanced feature space which uses multiple modalities improves the robustness of the model to reduction in training data. The plot\ref{fig:training_data} shows that the model is able to converge to a low training loss .  

\subsubsection{Corruption of text input}
\begin{figure}[htbp]
\centering
    \begin{subfigure}[b]{0.45\textwidth}
    \centering
    \includegraphics[width=\textwidth]{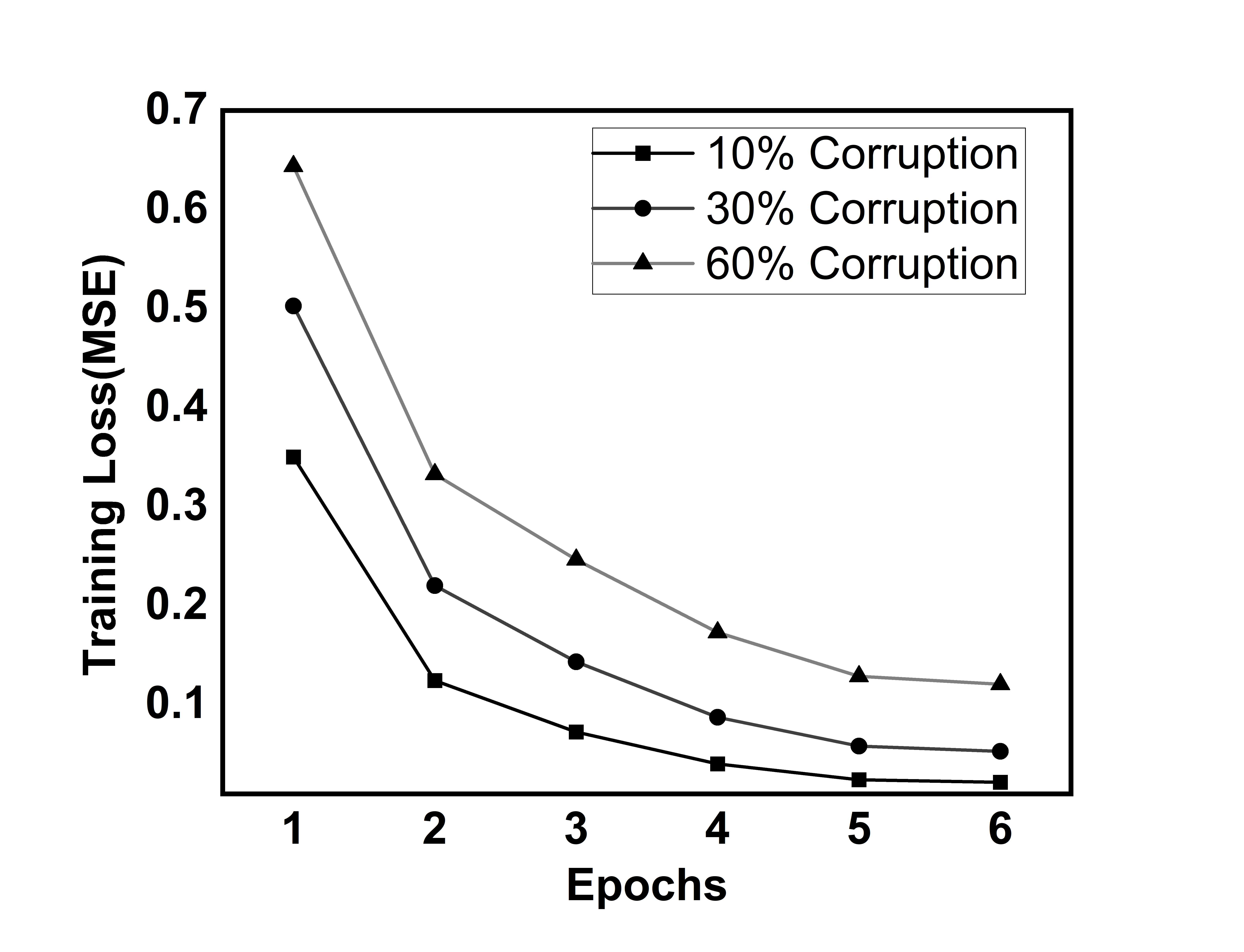}
    \caption{\label{fig:abs1}}
    \end{subfigure}
\quad
    \begin{subfigure}[b]{0.45\textwidth}
    \centering
    \includegraphics[width=\textwidth]{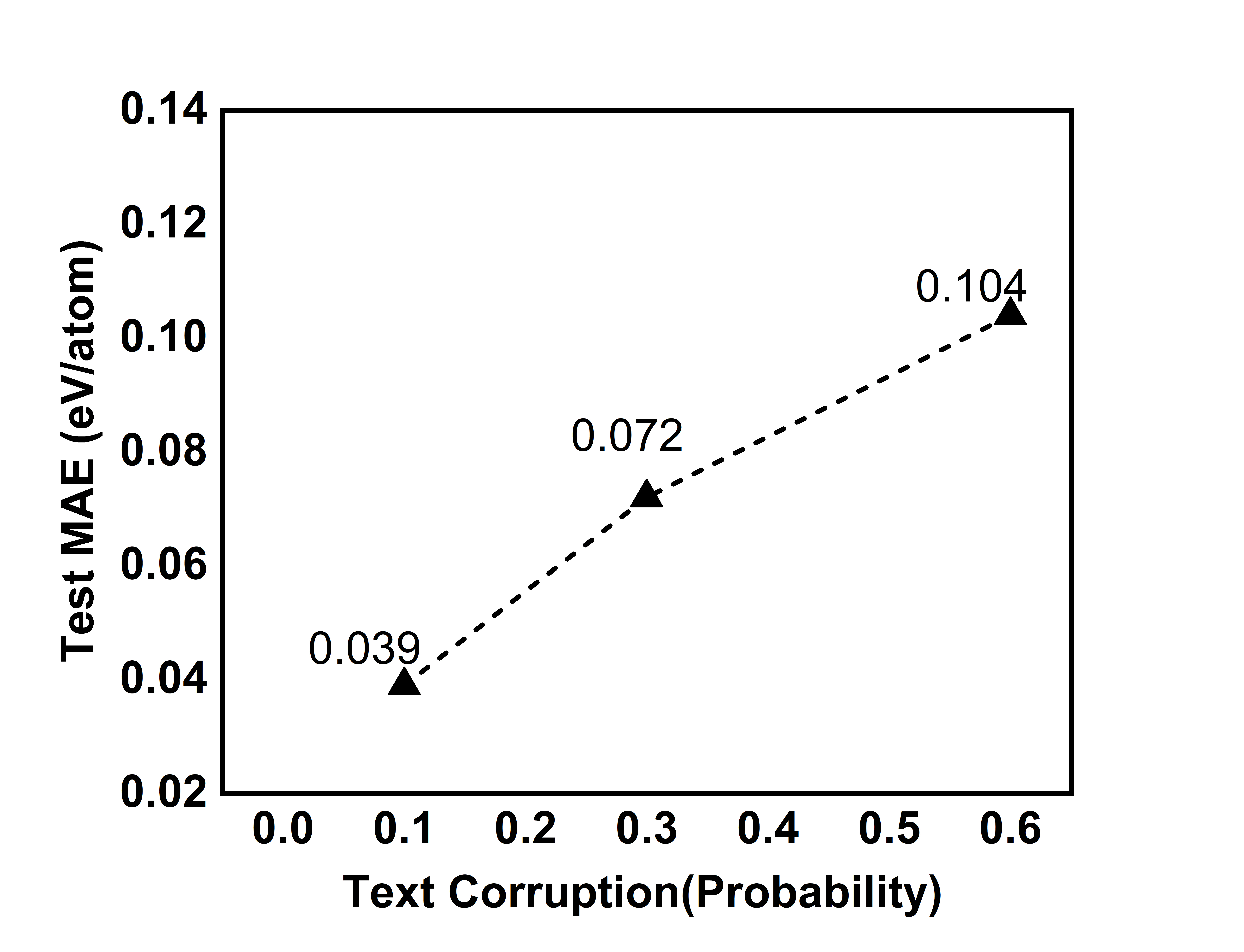}
    \caption{\label{fig:abs2}}
    \end{subfigure}
    \caption{The figure (\ref{fig:abs1}) and figure (\ref{fig:abs2}) shows the effect of the corruption of the input text on the training loss and the test loss respectively. The model performance deteriorates significantly with corruption in text.}
    \label{fig:ablation_corrupt_text}
\end{figure}

Corruption of text input remains a limitation of BERT models \citep{jin2020bert}. Moreover, Robocrsytallographer might lead to corrupted text output if there are aberrations in the CIF file \citep{ganose2019robocrystallographer}. Thus, we have studied the effect of different levels of text corruption on the performance of the model for predicting the formation energy per atom. For corrupting the text, we have deleted random characters, added random punctuations and performed random word substitutions. The level of corruption has been controlled by using the probability of corruption. There is a significant decrease in model performance as captured by the plot. The plot\ref{fig:ablation_corrupt_text} captures the degradation in model performance in training and inference with the increase in corruption of the text input.

\section{Conclusion}
This paper explores a multi-modal fusion model for predicting material properties. The Material Multi-Modal Fusion(\textbf{MatMMFuse}) model uses a multi-head cross attention based method for combining the embedding from graph neural network and a LLM model. The CGCNN model has been selected to encode the lattice structure as a graph encoding while, the SciBERT model has been used to encode the text descriptors. The SciBERT model already posses domain specific scientific knowledge which is helpful for generating meaningful embeddings. The enhanced feature space with the attention mechanism allows the model to selectively focus on key features from the structure aware graph embedding and the context aware embedding. The graph encoder focuses on local information while the text encoder is able to learn global information such as symmetry and space group. The results show that the proposed model is able to outperform both the plain vanilla versions of CGCNN and SciBERT models by 35\% and 68\% respectively for predicting the formation energy per atom. We observe an improvement for the Energy above Hull and the Fermi energy as well. Further, we observe an marginal improvement for the prediction of Band Gap which is aligned to the state of the art. Interestingly, we demonstrate that the zero shot performance of the model is better than the vanilla CGCNN and SciBERT models for cubic oxide perovskites, chalcogenide perovskites and a subset of the JARVIS datasets which is an important step for specialized uses cases. Analyses of the t-SNE plots show that our model is able to generate embeddings which have clear lobe-shaped decision boundaries and similar material properties are clustered together. Finally, we believe the ability of LLM models to use text based inputs for probing the underlying mechanism of the model to understand specific points of failure provides a tool to analyze the structure property relationships in crystalline solids.\newline
\textbf{Limitations and Future scope of work}: The model is unable to accurately predict the band gap for near zero values. A possible explanation might be the lack of experimental data. MatMMFuse has been designed to work only with CIF Files and thus, performance might be improved with grounding in experimental data. Importantly, quantum effects become more dominant at very small energy gaps. Thus, a possible area of improvement would be the explicit incorporation of quantum effects. Furthermore, we believe that additional modalities might lead to an improvement in the model performance. The cross attention operation scales quadratically with sequence length which makes it computationally expensive for long sequences. Also, it is possible for one modality to dominate the training process leading to imbalance. Thus, it would be interesting to explore alternate approaches for a balanced integration of modalities.

\section{Reproducibility} 

\textbf{Data Availability:} The Material project dataset analyzed during this study is available at https://next-gen.materialsproject.org/. The cubic oxide perovskite and chalcogenides datasets are available in the Computational Material Repository(https://cmr.fysik.dtu.dk). The JARVIS dataset is available at https://jarvis.nist.gov/.
\newline
\textbf{Code availability:} The code is publicly available at the github repository-
\url{https://github.com/AbhiroopBhattacharya/MatMMFuse}. 
Moreover, the pseudocode is also provided in the Appendix section. 

\section{Acknowledgments}
SGC thanks the Canada Research Chair and the NSERC Discovery programs for their support.

\section*{Author Contributions}
The concept and methodology were planned and done by AB and SGC. The first version of manuscript was written by AB. The manuscript was reviewed and commented by  SGC.


\newpage
\appendix
\section{Appendix}

\subsection{Dataset Description}
\subsubsection{Materials Project}
We leverage the widely used Materials Project dataset \citep{jain2013materials}. We focus on four important material properties: the formation energy per atom, the energy above the hull, the fermi energy and the Band Gap. The table \ref{tab:summary_stats} shows the distribution of the target variables. 

\begin{table}[htbp]
\centering
\caption{Summary Statistics for Target Variable for Materials Project Dataset}
\begin{tabular}{@{}ccccc@{}}\toprule
\multicolumn{4}{c}{\textbf{Target Variable Statistics}} \\
\cmidrule{2-5} 
& \multirow{2}{*}{\shortstack{Formation Energy \\[0.2em] (eV/atom)}} & \multirow{2}{*}{\shortstack{Fermi Energy \\[0.2em] (eV)}} & \multirow{2}{*}{\shortstack{Energy Above Convex Hull \\[0.2em] (eV/atom)}} & \multirow{2}{*}{\shortstack{Band Gap \\[0.2em] (eV)}} \\ 
& & & & \\ \midrule
Mean & -1.66 & 3.069 & 0.022 & 0.874 \\
Standard Deviation & 1.009 & 2.776 & 0.244 & 1.514\\
Range & [-11.86, 5.45] & [ -14.017, 19.41] & [0.00, 7.497] &[0.00, 17.891] \\
Median & -1.75 & 3.024 &0.00 & 0.00\\
\hline
\end{tabular}
\label{tab:summary_stats}
\end{table}

We use Robocrystallographer \citep{ganose2019robocrystallographer} to convert the crystal data encoded in CIF file into text format. The distribution of the generated text descriptions are given below in the table\ref{tab:summary_stats_text}
\begin{table}[htbp]
\centering
\caption{Summary Statistics for Text descriptions for Materials Project.}
\begin{tabular}{lc}\toprule
\multicolumn{2}{c}{\textbf{Text Description Statistics}}\\
\midrule
Average Length & 741.4 words \\
Standard Deviation & 1426.9 words \\
Range & [28, 49051] words \\
\hline
\end{tabular}
\label{tab:summary_stats_text}
\end{table}
\subsubsection{Cubic Oxide Perovskites}

We use the cubic oxide perovskite $ABO_3$ dataset from Computational material repository for evaluating the Zero shot performance of MatMMFuse. The summary statistics of the text descriptions are given in the table\ref{tab:summary_stats_text_abo3}.


\begin{table}[htbp]
\centering
\caption{Summary Statistics for Text descriptions for Cubic Oxide perovskites($AB0_3$)}
\begin{tabular}{lc}\toprule
\multicolumn{2}{c}{\textbf{Text Description Statistics}}\\
\midrule
Average Length & 136.6 words \\
Standard Deviation & 31.6 words \\
Range & [79, 239] words \\
\hline
\end{tabular}
\label{tab:summary_stats_text_abo3}
\end{table}

\subsubsection{Chalcogenides}
Chalcogenide perovskites have the form $AB(S,Se)_3$. We have used them for evaluating the zero shot performance of MatMMFuse. The summary statistics of the text descriptions are tabulated in the table\ref{tab:summary_stats_text_abs3}.

\begin{table}[htbp]
\centering
\caption{Summary Statistics for Text descriptions for Chalcogenide perovskites($ABS_3$,$ABSe_3$)}
\begin{tabular}{lc}\toprule
\multicolumn{2}{c}{\textbf{Text Description Statistics}}\\
\midrule
Average Length & 199.2 words \\
Standard Deviation & 103.2 words \\
Range & [63, 1641] words \\
\hline
\end{tabular}
\label{tab:summary_stats_text_abs3}
\end{table}

\subsection{Pseudocode for implementing proposed framework}

MatMMFuse can be implemented using the following pseudocode.
\begin{algorithm}
\caption{Fusion of Graph and Text Embeddings for Material Property Prediction}
\begin{algorithmic}[1]
\Require CIF file $\mathcal{C}$, Text Description $\mathcal{T}$, Property Label $y$, Pretrained GNN $\mathcal{G}$, Pretrained Transformer $\mathcal{B}$, Attention Combiner $\mathcal{A}$, Learning Rate $\eta$, Cosine Warmup $\lambda$
\Ensure Trained Model for Property Prediction

\State \textbf{Initialize} model parameters $\theta$
\State \textbf{Split} dataset into Train $(\mathcal{D}_{train})$, Validation $(\mathcal{D}_{val})$, and Test $(\mathcal{D}_{test})$

\For{each epoch in $1, \dots, N_{epochs}$}
    \For{each batch $(\mathcal{C}_i, \mathcal{T}_i, y_i)$ in $\mathcal{D}_{train}$}
        \State \textbf{Extract Graph Features:}
        \State \quad Construct crystal graph $G_i$ from CIF file $\mathcal{C}_i$
        \State \quad Compute graph embedding: $h_G = \mathcal{G}(G_i)$
        \State \quad Project embedding: $\tilde{h}_G = W_G h_G$

        \State \textbf{Extract Text Features:}
        \State \quad Tokenize text: $X_T = \text{Tokenizer}(\mathcal{T}_i)$
        \State \quad Compute transformer embedding: $h_T = \mathcal{B}(X_T)$
        \State \quad Pool embedding: $h_T = \text{Mean}(h_T)$
        \State \quad Project embedding: $\tilde{h}_T = W_T h_T$

        \State \textbf{Fuse Representations using Attention:}
        \State \quad Compute query: $Q = W_Q \tilde{h}_T$
        \State \quad Compute key: $K = W_K \tilde{h}_G$
        \State \quad Compute value: $V = W_V \tilde{h}_G$
        \State \quad Compute attention scores: $\alpha = \text{softmax} \left( \frac{QK^T}{\sqrt{d_k}} \right)$
        \State \quad Compute attended representation: $h_{fused} = \alpha V$
        \State \quad Apply residual connection: $h_{fused} = \text{LayerNorm}(h_{fused} + \tilde{h}_T)$

        \State \textbf{Predict Property:}
        \State \quad $y_{\text{pred}} = \sigma(W_o h_{fused})$

        \State \textbf{Compute Loss:}
        \State \quad $\mathcal{L} = \frac{1}{N} \sum_{i=1}^{N} (y_i - y_{\text{pred}})^2$

        \State \textbf{Optimize Parameters:}
        \State \quad Compute gradients: $\nabla_{\theta} \mathcal{L}$
        \State \quad Update parameters: $\theta \gets \theta - \eta \cdot \lambda(t) \cdot \nabla_{\theta} \mathcal{L}$

    \EndFor
    \State Evaluate on $\mathcal{D}_{val}$ and adjust learning rate $\eta$
\EndFor

\State \textbf{Test Model:} Evaluate on $\mathcal{D}_{test}$

\end{algorithmic}
\end{algorithm}


\begin{thebibliography}{35}
\providecommand{\natexlab}[1]{#1}
\providecommand{\url}[1]{\texttt{#1}}
\expandafter\ifx\csname urlstyle\endcsname\relax
  \providecommand{\doi}[1]{doi: #1}\else
  \providecommand{\doi}{doi: \begingroup \urlstyle{rm}\Url}\fi

\bibitem[Back et~al.(2019)Back, Yoon, Tian, Zhong, Tran, and Ulissi]{back2019convolutional}
Seoin Back, Junwoong Yoon, Nianhan Tian, Wen Zhong, Kevin Tran, and Zachary~W Ulissi.
\newblock Convolutional neural network of atomic surface structures to predict binding energies for high-throughput screening of catalysts.
\newblock \emph{The journal of physical chemistry letters}, 10\penalty0 (15):\penalty0 4401--4408, 2019.

\bibitem[Basera \& Bhattacharya(2022)Basera and Bhattacharya]{basera2022chalcogenide}
Pooja Basera and Saswata Bhattacharya.
\newblock Chalcogenide perovskites (abs3; a= ba, ca, sr; b= hf, sn): An emerging class of semiconductors for optoelectronics.
\newblock \emph{The Journal of Physical Chemistry Letters}, 13\penalty0 (28):\penalty0 6439--6446, 2022.

\bibitem[Behler(2011)]{behler2011atom}
J{\"o}rg Behler.
\newblock Atom-centered symmetry functions for constructing high-dimensional neural network potentials.
\newblock \emph{The Journal of chemical physics}, 134\penalty0 (7), 2011.

\bibitem[Beltagy et~al.(2019)Beltagy, Lo, and Cohan]{beltagy2019scibert}
Iz~Beltagy, Kyle Lo, and Arman Cohan.
\newblock Scibert: A pretrained language model for scientific text.
\newblock \emph{arXiv preprint arXiv:1903.10676}, 2019.

\bibitem[Chen et~al.(2019)Chen, Ye, Zuo, Zheng, and Ong]{chen2019graph}
Chi Chen, Weike Ye, Yunxing Zuo, Chen Zheng, and Shyue~Ping Ong.
\newblock Graph networks as a universal machine learning framework for molecules and crystals.
\newblock \emph{Chemistry of Materials}, 31\penalty0 (9):\penalty0 3564--3572, 2019.

\bibitem[Chen et~al.(2020)Chen, Zuo, Ye, Li, Deng, and Ong]{chen2020critical}
Chi Chen, Yunxing Zuo, Weike Ye, Xiangguo Li, Zhi Deng, and Shyue~Ping Ong.
\newblock A critical review of machine learning of energy materials.
\newblock \emph{Advanced Energy Materials}, 10\penalty0 (8):\penalty0 1903242, 2020.

\bibitem[Choudhary et~al.(2020)Choudhary, Garrity, Reid, DeCost, Biacchi, Hight~Walker, Trautt, Hattrick-Simpers, Kusne, Centrone, et~al.]{choudhary2020joint}
Kamal Choudhary, Kevin~F Garrity, Andrew~CE Reid, Brian DeCost, Adam~J Biacchi, Angela~R Hight~Walker, Zachary Trautt, Jason Hattrick-Simpers, A~Gilad Kusne, Andrea Centrone, et~al.
\newblock The joint automated repository for various integrated simulations (jarvis) for data-driven materials design.
\newblock \emph{npj computational materials}, 6\penalty0 (1):\penalty0 173, 2020.

\bibitem[Das et~al.(2023)Das, Goyal, Lee, Bhattacharjee, and Ganguly]{das2023crysmmnet}
Kishalay Das, Pawan Goyal, Seung-Cheol Lee, Satadeep Bhattacharjee, and Niloy Ganguly.
\newblock Crysmmnet: multimodal representation for crystal property prediction.
\newblock In \emph{Uncertainty in Artificial Intelligence}, pp.\  507--517. PMLR, 2023.

\bibitem[De et~al.(2016)De, Bart{\'o}k, Cs{\'a}nyi, and Ceriotti]{de2016comparing}
Sandip De, Albert~P Bart{\'o}k, G{\'a}bor Cs{\'a}nyi, and Michele Ceriotti.
\newblock Comparing molecules and solids across structural and alchemical space.
\newblock \emph{Physical Chemistry Chemical Physics}, 18\penalty0 (20):\penalty0 13754--13769, 2016.

\bibitem[Draxl \& Scheffler(2019)Draxl and Scheffler]{draxl2019nomad}
Claudia Draxl and Matthias Scheffler.
\newblock The nomad laboratory: from data sharing to artificial intelligence.
\newblock \emph{Journal of Physics: Materials}, 2\penalty0 (3):\penalty0 036001, 2019.

\bibitem[Faber et~al.(2015)Faber, Lindmaa, Von~Lilienfeld, and Armiento]{faber2015crystal}
Felix Faber, Alexander Lindmaa, O~Anatole Von~Lilienfeld, and Rickard Armiento.
\newblock Crystal structure representations for machine learning models of formation energies.
\newblock \emph{International Journal of Quantum Chemistry}, 115\penalty0 (16):\penalty0 1094--1101, 2015.

\bibitem[Fung et~al.(2021)Fung, Zhang, Juarez, and Sumpter]{fung2021benchmarking}
Victor Fung, Jiaxin Zhang, Eric Juarez, and Bobby~G Sumpter.
\newblock Benchmarking graph neural networks for materials chemistry.
\newblock \emph{npj Computational Materials}, 7\penalty0 (1):\penalty0 84, 2021.

\bibitem[Ganose \& Jain(2019)Ganose and Jain]{ganose2019robocrystallographer}
Alex~M Ganose and Anubhav Jain.
\newblock Robocrystallographer: automated crystal structure text descriptions and analysis.
\newblock \emph{MRS Communications}, 9\penalty0 (3):\penalty0 874--881, 2019.

\bibitem[Gori et~al.(2005)Gori, Monfardini, and Scarselli]{gori2005new}
Marco Gori, Gabriele Monfardini, and Franco Scarselli.
\newblock A new model for learning in graph domains.
\newblock In \emph{Proceedings. 2005 IEEE International Joint Conference on Neural Networks, 2005.}, volume~2, pp.\  729--734. IEEE, 2005.

\bibitem[Ishihara(2009)]{ishihara2009structure}
Tatsumi Ishihara.
\newblock Structure and properties of perovskite oxides.
\newblock \emph{Perovskite Oxide for Solid Oxide Fuel Cells}, pp.\  1--16, 2009.

\bibitem[Jablonka et~al.(2023)Jablonka, Ai, Al-Feghali, Badhwar, Bocarsly, Bran, Bringuier, Brinson, Choudhary, Circi, et~al.]{jablonka202314}
Kevin~Maik Jablonka, Qianxiang Ai, Alexander Al-Feghali, Shruti Badhwar, Joshua~D Bocarsly, Andres~M Bran, Stefan Bringuier, L~Catherine Brinson, Kamal Choudhary, Defne Circi, et~al.
\newblock 14 examples of how llms can transform materials science and chemistry: a reflection on a large language model hackathon.
\newblock \emph{Digital Discovery}, 2\penalty0 (5):\penalty0 1233--1250, 2023.

\bibitem[Jain et~al.(2013)Jain, Ong, Hautier, Chen, Richards, Dacek, Cholia, Gunter, Skinner, Ceder, et~al.]{jain2013materials}
Anubhav Jain, Shyue~Ping Ong, Geoffroy Hautier, Wei Chen, William~Davidson Richards, Stephen Dacek, Shreyas Cholia, Dan Gunter, David Skinner, Gerbrand Ceder, et~al.
\newblock The materials project: A materials genome approach to accelerating materials innovation, apl mater.
\newblock \emph{Applied Physics Letter}, 2013.

\bibitem[Jin et~al.(2020)Jin, Jin, Zhou, and Szolovits]{jin2020bert}
Di~Jin, Zhijing Jin, Joey~Tianyi Zhou, and Peter Szolovits.
\newblock Is bert really robust? a strong baseline for natural language attack on text classification and entailment.
\newblock In \emph{Proceedings of the AAAI conference on artificial intelligence}, volume~34, pp.\  8018--8025, 2020.

\bibitem[Lan et~al.(2019)Lan, Chen, Goodman, Gimpel, Sharma, and Soricut]{lan2019albert}
Zhenzhong Lan, Mingda Chen, Sebastian Goodman, Kevin Gimpel, Piyush Sharma, and Radu Soricut.
\newblock Albert: A lite bert for self-supervised learning of language representations.
\newblock \emph{arXiv preprint arXiv:1909.11942}, 2019.

\bibitem[Lee et~al.(2025)Lee, Park, Yang, Lim, and Han]{lee2025cast}
Jaewan Lee, Changyoung Park, Hongjun Yang, Sungbin Lim, and Sehui Han.
\newblock Cast: Cross attention based multimodal fusion of structure and text for materials property prediction.
\newblock \emph{arXiv preprint arXiv:2502.06836}, 2025.

\bibitem[Li et~al.(2024)Li, Shomer, Mao, Zeng, Ma, Shah, Tang, and Yin]{li2024evaluating}
Juanhui Li, Harry Shomer, Haitao Mao, Shenglai Zeng, Yao Ma, Neil Shah, Jiliang Tang, and Dawei Yin.
\newblock Evaluating graph neural networks for link prediction: Current pitfalls and new benchmarking.
\newblock \emph{Advances in Neural Information Processing Systems}, 36, 2024.

\bibitem[Li et~al.(2025)Li, Gupta, Kilic, Choudhary, Wines, Liao, Choudhary, and Agrawal]{li2025hybrid}
Youjia Li, Vishu Gupta, Muhammed Nur~Talha Kilic, Kamal Choudhary, Daniel Wines, Wei-keng Liao, Alok Choudhary, and Ankit Agrawal.
\newblock Hybrid-llm-gnn: integrating large language models and graph neural networks for enhanced materials property prediction.
\newblock \emph{Digital Discovery}, 2025.

\bibitem[Masala et~al.(2020)Masala, Ruseti, and Dascalu]{masala2020robert}
Mihai Masala, Stefan Ruseti, and Mihai Dascalu.
\newblock Robert--a romanian bert model.
\newblock In \emph{Proceedings of the 28th International Conference on Computational Linguistics}, pp.\  6626--6637, 2020.

\bibitem[Ock et~al.(2024)Ock, Montoya, Schweigert, Hung, Suram, and Ye]{ock2024unimat}
Janghoon Ock, Joseph Montoya, Daniel Schweigert, Linda Hung, Santosh~K Suram, and Weike Ye.
\newblock Unimat: Unifying materials embeddings through multi-modal learning.
\newblock \emph{arXiv preprint arXiv:2411.08664}, 2024.

\bibitem[Palizhati et~al.(2019)Palizhati, Zhong, Tran, Back, and Ulissi]{palizhati2019toward}
Aini Palizhati, Wen Zhong, Kevin Tran, Seoin Back, and Zachary~W Ulissi.
\newblock Toward predicting intermetallics surface properties with high-throughput dft and convolutional neural networks.
\newblock \emph{Journal of chemical information and modeling}, 59\penalty0 (11):\penalty0 4742--4749, 2019.

\bibitem[Paszke et~al.(2017)Paszke, Gross, Chintala, Chanan, Yang, DeVito, Lin, Desmaison, Antiga, and Lerer]{paszke2017automatic}
Adam Paszke, Sam Gross, Soumith Chintala, Gregory Chanan, Edward Yang, Zachary DeVito, Zeming Lin, Alban Desmaison, Luca Antiga, and Adam Lerer.
\newblock Automatic differentiation in pytorch.
\newblock \emph{Neural Information Processing Systems}, 2017.

\bibitem[Saal et~al.(2013)Saal, Kirklin, Aykol, Meredig, and Wolverton]{saal2013materials}
James~E Saal, Scott Kirklin, Muratahan Aykol, Bryce Meredig, and Christopher Wolverton.
\newblock Materials design and discovery with high-throughput density functional theory: the open quantum materials database (oqmd).
\newblock \emph{Jom}, 65:\penalty0 1501--1509, 2013.

\bibitem[Sanh et~al.(2019)Sanh, Debut, Chaumond, and Wolf]{sanh2019distilbert}
Victor Sanh, Lysandre Debut, Julien Chaumond, and Thomas Wolf.
\newblock Distilbert, a distilled version of bert: smaller, faster, cheaper and lighter.
\newblock \emph{arXiv preprint arXiv:1910.01108}, 2019.

\bibitem[Scarselli et~al.(2008)Scarselli, Gori, Tsoi, Hagenbuchner, and Monfardini]{scarselli2008graph}
Franco Scarselli, Marco Gori, Ah~Chung Tsoi, Markus Hagenbuchner, and Gabriele Monfardini.
\newblock The graph neural network model.
\newblock \emph{IEEE transactions on neural networks}, 20\penalty0 (1):\penalty0 61--80, 2008.

\bibitem[Schmidt et~al.(2019)Schmidt, Marques, Botti, and Marques]{schmidt2019recent}
Jonathan Schmidt, M{\'a}rio~RG Marques, Silvana Botti, and Miguel~AL Marques.
\newblock Recent advances and applications of machine learning in solid-state materials science.
\newblock \emph{npj Computational Materials}, 5\penalty0 (1):\penalty0 83, 2019.

\bibitem[Sch{\"u}tt et~al.(2018)Sch{\"u}tt, Sauceda, Kindermans, Tkatchenko, and M{\"u}ller]{schutt2018schnet}
Kristof~T Sch{\"u}tt, Huziel~E Sauceda, P-J Kindermans, Alexandre Tkatchenko, and K-R M{\"u}ller.
\newblock Schnet--a deep learning architecture for molecules and materials.
\newblock \emph{The Journal of Chemical Physics}, 148\penalty0 (24), 2018.

\bibitem[Sergio \& Lee(2021)Sergio and Lee]{sergio2021stacked}
Gwenaelle~Cunha Sergio and Minho Lee.
\newblock Stacked debert: All attention in incomplete data for text classification.
\newblock \emph{Neural Networks}, 136:\penalty0 87--96, 2021.

\bibitem[Van~der Maaten \& Hinton(2008)Van~der Maaten and Hinton]{van2008visualizing}
Laurens Van~der Maaten and Geoffrey Hinton.
\newblock Visualizing data using t-sne.
\newblock \emph{Journal of machine learning research}, 9\penalty0 (11), 2008.

\bibitem[Xie \& Grossman(2018)Xie and Grossman]{xie2018crystal}
Tian Xie and Jeffrey~C Grossman.
\newblock Crystal graph convolutional neural networks for an accurate and interpretable prediction of material properties.
\newblock \emph{Physical review letters}, 120\penalty0 (14):\penalty0 145301, 2018.

\bibitem[Zhuo et~al.(2018)Zhuo, Mansouri~Tehrani, and Brgoch]{zhuo2018predicting}
Ya~Zhuo, Aria Mansouri~Tehrani, and Jakoah Brgoch.
\newblock Predicting the band gaps of inorganic solids by machine learning.
\newblock \emph{The journal of physical chemistry letters}, 9\penalty0 (7):\penalty0 1668--1673, 2018.

\end{thebibliography}
\end{document}